\documentclass[conference]{IEEEtran}
\IEEEoverridecommandlockouts
\usepackage{graphicx} 
\usepackage{amsmath}
\usepackage{hyperref}
\usepackage{comment}
\usepackage{caption}
\usepackage{subcaption}
\usepackage[dvipsnames]{xcolor}
\usepackage{xspace}
\usepackage[latin1]{inputenc}

\newif\ifshowcomments
\showcommentsfalse

\usepackage{xcolor}

\title{IEEEIV-CARLA-Real-World-Traffic-Data}
\author{Anonymous}
\date{November 2025}

\begin{document}

\title{Incorporating Ephemeral Traffic Waves in\\ A Data-Driven Framework for\\ Microsimulation in CARLA}

\author{\IEEEauthorblockN{Alex Richardson}
\IEEEauthorblockA{\textit{Computer Science} \\
\textit{Vanderbilt University}\\
Nashville, TN, USA  \\
william.a.richardson@vanderbilt.edu}
\and
\IEEEauthorblockN{Azhar Hasan}
\IEEEauthorblockA{\textit{Computer Science} \\
\textit{Vanderbilt University}\\
Nashville, TN, USA  \\
}
\and
\IEEEauthorblockN{Gabor Karsai}
\IEEEauthorblockA{\textit{Computer Science} \\
\textit{Vanderbilt University}\\
Nashville, TN, USA  \\
}
\and
\IEEEauthorblockN{Jonathan Sprinkle}
\IEEEauthorblockA{\textit{Computer Science} \\
\textit{Vanderbilt University}\\
Nashville, TN, USA  \\
jonathan.sprinkle@vanderbilt.edu}
}

\maketitle

\begin{abstract}
This paper introduces a data-driven traffic microsimulation framework in CARLA that reconstructs real-world wave dynamics using high-fidelity time-space data from the I-24 MOTION testbed. 
Calibration of road networks in microsimulators to reproduce ephemeral phenomena such as traffic waves for large-scale simulation is a process that is fraught with challenges. 
This work reconsiders the existence of the traffic state data as boundary conditions on an ego vehicle moving through previously recorded traffic data, rather than reproducing those traffic phenomena in a calibrated microsim.
Our approach is to autogenerate a 1 mile highway segment corresponding to I-24, and use the I-24 data to power a cosimulation module that injects traffic information into the simulation. 
The CARLA and cosimulation simulations are centered around an ego vehicle sampled from the empirical data, with autogeneration of "visible" traffic within the longitudinal range of the ego vehicle. 
Boundary control beyond these visible ranges is achieved using ghost cells behind (upstream) and ahead (downstream) of the ego vehicle. 
Unlike prior simulation work that focuses on local car-following behavior or abstract geometries, our framework targets full time-space diagram fidelity as the validation objective. 
Leveraging CARLA's rich sensor suite and configurable vehicle dynamics, we simulate wave formation and dissipation in both low-congestion and high-congestion scenarios for qualitative analysis. 
The resulting emergent behavior closely mirrors that of real traffic, providing a novel cosimulation framework for evaluating traffic control strategies, perception-driven autonomy, and future deployment of wave mitigation solutions. 
Our work bridges microscopic modeling with physical experimental data, enabling the first perceptually realistic, boundary-driven simulation of empirical traffic wave phenomena in CARLA.
\end{abstract}


\section{Introduction}
Stop-and-go traffic waves are continuing to be an object of study for advancement in traffic flow theory, and in cyber-physical transportation systems as a means to understand how they can be controlled. While the emergence and propagation of these waves are well-documented in empirical studies and increasingly targeted by control strategies such as adaptive cruise control (ACC), and Vehicle-to-Vehicle (V2V) coordination, there remains a fundamental gap in evaluating the real-world fidelity of mitigation techniques under realistic traffic conditions.

To date, many traffic microsimulations used to study stop-and-go waves rely on abstract road geometries \cite{flow_wu_2021}, periodic boundary conditions (like ring roads), or synthetic traffic generation. While useful for theoretical exploration or controller training, such environments fall short in replicating the spatiotemporal complexity of real traffic dynamics. Moreover, existing simulation platforms rarely account for the full physical and perceptual realism required to test Autonomous Vehicle (AV) systems under traffic wave conditions.

In this work, we introduce a data-driven, high-fidelity traffic microsimulation in CARLA that reconstructs emergent traffic wave conditions for the ego AV, using empirical data derived from the I-24 MOTION testbed. Our simulation models a 1 mile segment of the I-24 freeway in Tennessee---a corridor that captures vehicle trajectory data at high resolution. The simulation spawns a single ego vehicle from the empirical data at a selected timestamp, longitudinal position, and lane. Surrounding traffic within a longitudinal ``visible'' range from the ego vehicle is autogenerated and removed via a cosimulation module. The cosimulation module also tracks traffic trajectories in a ``ghost cell'' region behind and in front of this visible range, which serves as boundary control for the ``visible'' and ego vehicles. This architecture enables dynamically generated vehicles to enter and exit the area around the ego vehicle in accordance with real-world density profiles extracted from time-space diagrams.

Unlike previous studies that focus on local car-following behavior or theoretical controller performance, our simulation framework targets the fidelity of the full time-space diagram as the primary validation objective. Using CARLA, we simulate traffic dynamics under physically plausible conditions under 2 selected scenarios---a low-congestion and high-congestion scenario---and qualitatively evaluate the simulated ego vehicle's behavior against the empirical data.

\subsection*{Contributions}
\begin{enumerate}
    \item We develop a CARLA-based traffic microsimulation aligned with a real 1 mile I-24 highway segment, integrating empirical data and road geometry. The software that executes this coordination between CARLA and the I-24 Motion data is available at \cite{thiscode}.
    \item We demonstrate the first empirically sourced simulation of phantom jam behavior in CARLA, enabling future controller testing, autonomy validation, and policy planning.
    \item We propose a ghost cell boundary control method for CARLA that enables realistic vehicle injection and exit based on time-space density profiles from real traffic.
    \item We propose time-space diagram fidelity as a core metric for validating emergent traffic behavior in full 3D AV simulation against real-world data.
\end{enumerate}

\section{Related Work}

\subsection{Stop-and-go Traffic Wave Data and Theory}
Extensive research has characterized the behavior of stop-and-go waves empirically in real world highways \cite{traffic_treiber_2012,congested_treiber_2000} and in artificial real-world periodic boundary conditions \cite{traffic_sugiyama_2008}. In part due to the prohibitive cost of obtaining empirical data in the first decades of the field, hydrodynamic and car-following mathematical theories arose to characterize real-world behavior \cite{congested_treiber_2000,behavioural_gipps_1981,dynamical_bando_1995}. With the advent of large-scale empirical data \cite{gloudemans2023i24motion}, statistical tools have also arisen to analyze real-world waves \cite{scalable_ji_2024}.

\subsection{Control of Stop-and-Go Waves}
Many techniques for mitigating stop-and-go waves exist and are in active development. For localized vehicular control, the first major technique emerged in the 1990s in the form of noncooperative Adaptive Cruise Control \cite{autonomous_ioannou_1993}. 
While the results of that theory demonstrated approaches that would reduce amplification of disturbances, experimental results showed that manufacturers often were unable to directly implement that theory, resulting in string-unstable commercial systems \cite{commercially_gunter_2019}. Recent research has emerged on various forms of cooperative vehicle stop-and-go wave mitigation, as part of large-scale field experiments
\cite{traffic_lee_2025}.
\subsection{Real World Traffic Flow Data Collection}
For decades, microscopic empirical traffic flow data was prohibitive to collect at scale, which hindered development of traffic flow theory and mitigation techniques. Several approaches have converged to ameliorate this in the last decade.

Onboard vehicle data collection by public researchers has indicated that large-scale onboard telemetric data collection can be performed by small-scale research teams, using CAN data collection \cite{libpanda_bunting_2021,ros_elmadani_2021}.

Video-based data collection with the advent of computer vision-based tracking has also surged in popularity. Drone-based datasets have become available \cite{highd_krajewski_2018}, and recently, I-24 MOTION \cite{i24_gloudemans_2023} demonstrated that large-scale microscopic traffic flow data collection is possible from fixed camera poles, with public datasets containing millions of trajectories.

\subsection{Bird's Eye View Traffic Simulations}
Simulations of stop-and-go wave mitigation, for wave analysis, and mitigation technique development, has largely taken the form of Bird's Eye View simulations 
\cite{intelligent_treiber_2017, survey_di_2021,
microscopic_lopez_2018,
benchmarks_vinitsky_2018,
flow_wu_2017,flow_wu_2021,
parsimonious_laval_2014,
recurrent_zhou_2017,
neural_panwai_2007,
enhanced_sharath_2020} 
and macroscopic fluid-dynamic based simulations \cite{traffic_zehe_2015}.

However, in recent years, fully immersive 3D Autonomous Vehicle simulators, such as CARLA \cite{carla_dosovitskiy_2017}, have become broadly available and feasible to use at scale with hundreds of agents. These offer a diverse selection of potential vehicle and infrastructure models, along with comprehensive meteorological effects. They have been extensively employed in deep RL for end-to-end AV solutions \cite{survey_chao_2020,multiagent_bhattacharyya_2018,carla_osiski_2020,safety_xu_2025}. For traffic stop-and-go wave mitigation, there is great potential for their use, such as in software testing and digital twins.

\section{CARLA Cosimulation Framework}
This section details the structure and implementation of the cosimulation architecture built to interface with CARLA for replicating real-world traffic wave dynamics observed on the I-24 freeway. The simulation combines empirical data, realistic road geometry, calibrated vehicle control models, and a novel ghost cell mechanism to recreate emergent stop-and-go phenomena with high physical and behavioral fidelity. CARLA, and the vehicles therein, operate according to CARLA's system design and logic, and are managed by the stock CARLA traffic manager. However, CARLA is partnered with a custom cosimulator module that manages the overall simulation. In this simulation framework, there is an ego vehicle, the visible vehicles -     those within a longitudinal distance threshold of the ego vehicle, and the ghost vehicles - the vehicles in the ghost cell regions behind and in front of the ego vehicle.

\subsection{I-24 Environment}
The section of I-24 we use in this simulation study is a 1 mile section, from the 60.6 mile marker, to the 61.6 mile marker. This section of I-24 consists of two straight away roads, one heading southeast away from the Nashville metro area (the ``Eastbound'' section), and the other heading northwest towards the Nashville metro area (the ``Westbound'' section). During the morning rush hour, the Westbound section is in high demand, whereas in the afternoon, the Eastbound section experiences the corresponding surge. Both roads have 4 lanes, with the left most lane being a High-Occupancy Vehicle (HOV) lane.

This multi-lane highway section has minimal curvature, and no interchanges or ramps, ensuring that emergent traffic phenomena are primarily impacted by vehicle behavior and vehicle inflow/outflow constraints.

\begin{figure*}
    \centering
    \includegraphics[width=0.75\linewidth]{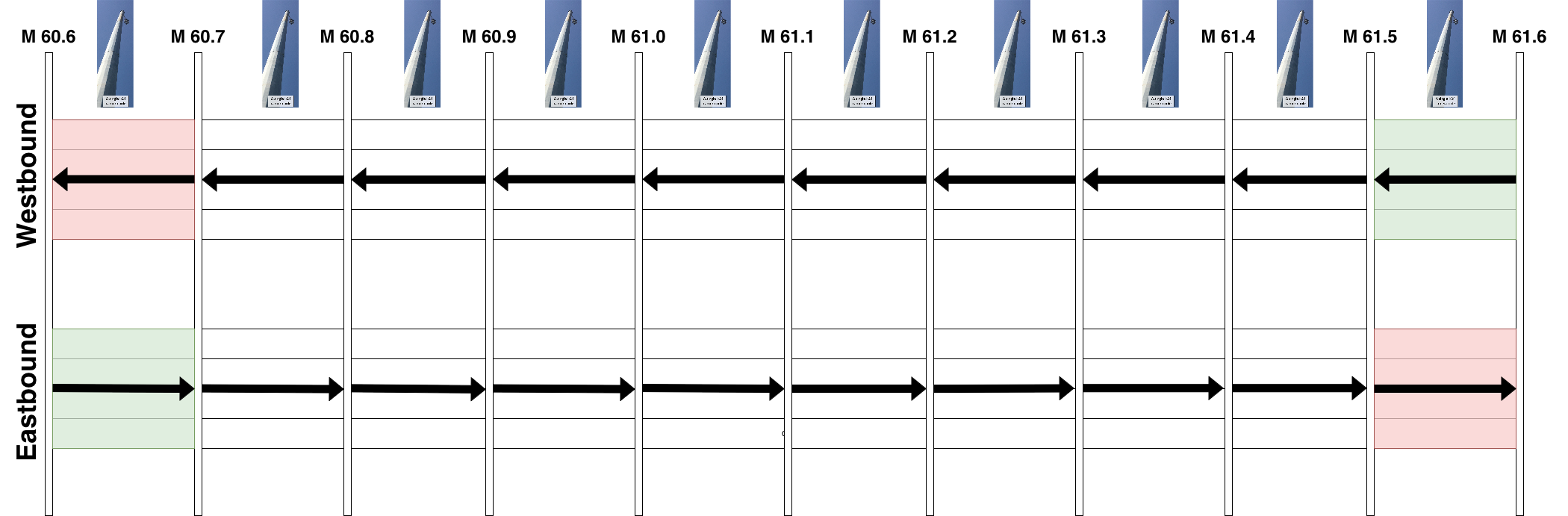}
    \caption{The I-24 60.6 mile to 61.6 mile region shown as a simplified diagram, with each segment at 0.1 mile in length demarcated. Cameras are shown at approximate spacing, with the green segments showing the beginning of each road, and the red segments showing the end of each road.}
    \label{fig:carla_map}
\end{figure*}

\subsection{Road and Environment Generation}
To generate the roads and terrain, we use a custom in-house map generator to procedurally create a basic terrain mesh, along with the 1 mile I-24 highway section. This generates an elevation-aware road topography, along with all the relevant lanes.

Road surface friction directly uses the stock CARLA road surface friction parameters. Lane markings and signs are not generated for the purposes of this simulation study.

Meteorological conditions consist of CARLA's default "high noon" setting, with no precipitation.

\begin{figure*}
    \centering
    \includegraphics[width=0.8\linewidth]{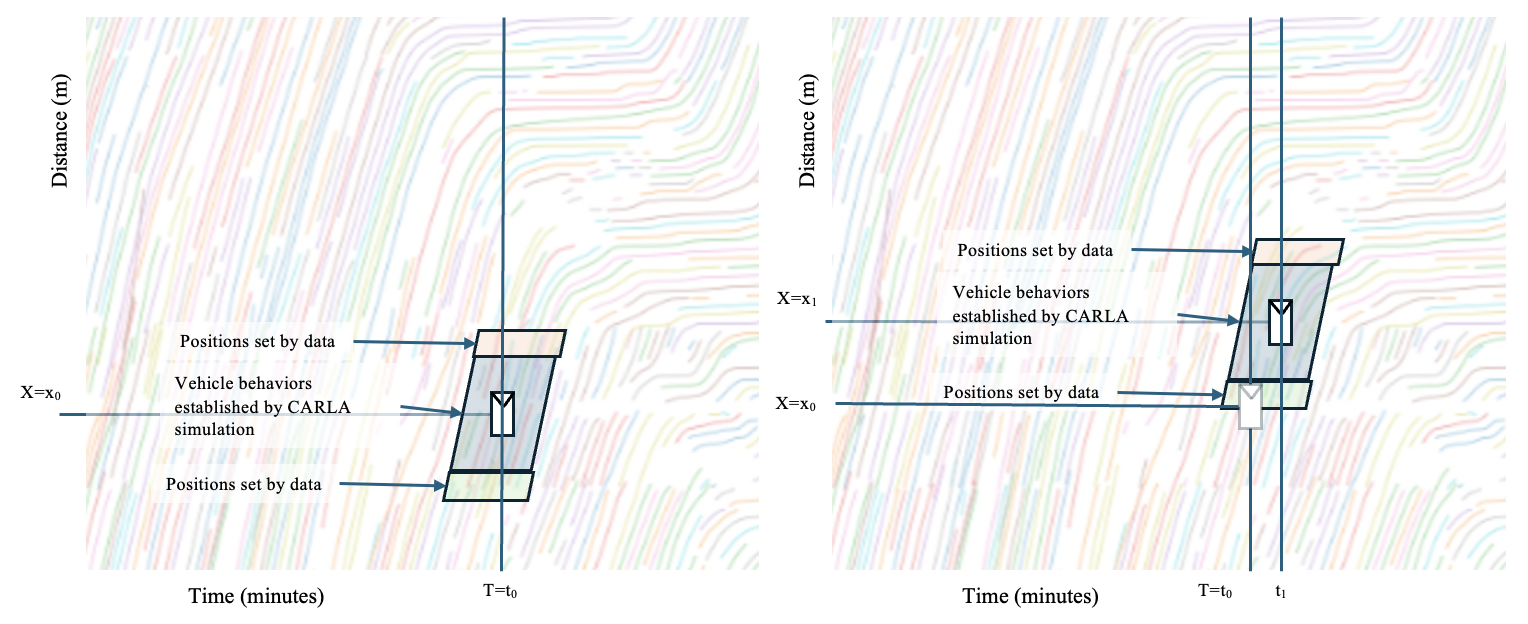}
    \caption{Two time-space diagrams with overlay of the ego vehicle and its surroundings at two time, $t_0$ and $t_1$. The direction of travel is ``up'' in the figure, and time advances to the right. (Left) at time $T=t_0$. Downstream cars are placed in their exact positions and given their velocity directly based on the representative traffic data. The simulated cars in CARLA are in the the immediate surroundings of the ego vehicle, and managed by its traffic control framework. Upstream cars (lower in the figure) are positioned directly by data. 
    (Right) at time $T=t_1$ the ego car and its surroundings have moved downstream. The car has now encountered a traffic wave, which means that additional cars will begin to appear in the local CARLA framework as the surroundings pick those cars up directly from the positional information gained by the measured data. 
    }
    \label{fig:timespace-overview}
\end{figure*}

\subsection{Cosimulator Design}
\subsubsection{Vehicle State}
Each vehicle instantaneous state with the identifier $id$ is designated as $\stackrel{id}{v}$. Each vehicle has the corresponding internal state:
\begin{itemize}
    \item $\stackrel{id}{v}_t$: The vehicle's current unix timestamp in seconds.
    \item $\stackrel{id}{v}_x$: The vehicle's current longitudinal position on the road in meters as described in Figure \ref{fig:carla_map}.
    \item $\stackrel{id}{v}_y$: The vehicle's current lateral position on the road in meters as described in Figure \ref{fig:carla_map}. 
    \item $\stackrel{id}{v}_{l}$: The vehicle's current lane -  in this work, $l \in \{-1, -2, -3, -4\}$: $-1$ corresponds to the leftmost lane (which is HOV), and $-4$ corresponds to the rightmost lane. The identifiers descend in a left to right order.
    \item $\stackrel{id}{v}_{\text{vel}}$: The vehicle's current longitudinal velocity in $\frac{m}{s}$.
\end{itemize}
\subsubsection{Configuration}
There are several configuration options the cosimulator has:
\begin{itemize}
    \item Episode time-length ($t_{max}$): The total time length of the simulation, in seconds. By default, it is $30$ seconds.
    \item Time-step size ($\Delta t$): This is the size, in seconds, of the time-steps of both the CARLA simulation and the cosimulation module. By default, it is $0.01$ seconds.
    \item Default desired speed ($\text{vel}_{\text{default}}$): The default desired speed in $\frac{m}{s}$ for the ego and visible vehicles in CARLA.
    \item Desired initial road ($r$): This is the selected roadway we want our vehicle to be simulated on. In this work, $r \in \{\text{Westbound}, \text{Eastbound}\}$ - it is either the Westbound road, or the Eastbound road.
    \item Visible window ($w_{visible}$): This is the longitudinal window, in meters, for the visible vehicles. By default, it is $150$ meters. This extends from the ego vehicle's longitudinal location ($\stackrel{ego}{v}_{x}$) both backwards and forwards with respect to the direction of travel. This means the bounds of the $v_x$ of any visible vehicle are:
    \[\stackrel{ego}{v}_{x} - w_{visible} \le v_x \le \stackrel{ego}{v}_{x} + w_{visible}\]
    \item Ghost window ($w_{ghost}$): This is the longitudinal window, in meters, for the ghost vehicles. By default, this is $50$ meters. These are placed immediately in front and behind of the visible window. 
    
    Hence, the bounds of the $v_x$ of any ghost vehicle are:
    \[\stackrel{ego}{v}_{x} - w_{visible} - w_{ghost} \le v_x \le \stackrel{ego}{v}_{x} -w_{visible}\]
    or:
    \[\stackrel{ego}{v}_{x} + w_{visible} \le v_x \le \stackrel{ego}{v}_{x} + w_{visible} + w_{ghost}\]
    \item Desired initial lane, x, and timestamp ($\hat{\text{ego}}_{l}$, $\hat{\text{ego}}_{x}$, $\hat{\text{ego}}_{t}$): This is the lane, longitudinal position, and timestamp we desire for our selected ego vehicle in the initialization phase.
\end{itemize}

\subsubsection{Simulation State}
The simulation's internal state consists of the following:
\begin{itemize}
    \item Current timestamp ($t$): The current unix timestamp, in seconds, of the simulation.
    \item Timestamp origin ($t_{0}$): This is the first unix timestamp of the simulation. Corresponds directly to the original timestamp of the selected ego vehicle from the empirical data.
    \item Ego vehicle ($\stackrel{ego}{v}$): The ego vehicle of the simulation.
    \item Visible vehicle set ($V_{\text{visible}}$: The current set of $\stackrel{id}{v}$ that are currently in the visible range, spawned in CARLA, and behaviorally largely managed by the Traffic Manager (except for desired velocity).
    \item Ghost vehicle set ($V_{\text{ghost}}$: The current set of $\stackrel{id}{v}$ that are currently in the ghost cells, and updated by reloading them from the I-24 empirical data at each loop (see below).
    \item I-24 Empirical Dataset ($V_{\text{empirical}}$): This is the total set of timestamped instantaneous vehicle state data, corresponding in format to $\stackrel{id}{v}$, that is available. We use this to spawn the ego/visible/ghost vehicles in the first timestep, and update the ghost region as the simulation advances.
\end{itemize}
\subsubsection{CARLA Vehicles}
The ego vehicle and the visible vehicles are the only vehicles that are ever spawned and operated within CARLA itself - ghost vehicles are never created. The vehicle type for each vehicle in CARLA is randomly selected. All are controlled by CARLA's Traffic Manager. When spawned, each is given an instantaneous velocity by an appropriate impulse force.
\subsubsection{Simulation Control Flow}
The simulation performs the following steps:
\begin{enumerate}
    \item Initialize $\stackrel{ego}{v}$ to the spatiotemporally closest vehicle available in $V_{\text{empirical}}$. Set the timestamp origin $t_{0}$ and current timestamp $t$ to $\stackrel{ego}{v}_t$.
    \item Load into $V_{visible}$ and ghost vehicles $V_{ghost}$ all eligible vehicles from $V_{empirical}$ at timestamp $t$ that are not $\stackrel{ego}{v}$. Spawn the ego/visible vehicles into CARLA as appropriate.
    \item Update the CARLA Traffic Manager's desired velocity for each ego and visible vehicle. Ego vehicle is maintained at the default speed $\text{vel}_{\text{default}}$. Visible vehicles behind the ego vehicle are given the same speed. For visible vehicles ahead of the ego vehicle, it is more complex. If they are not the lead vehicle for their lane, their desired speed is the default speed. If they are the lead vehicle for their lane of visible vehicles, their desired velocity corresponds to the ahead ghost vehicle. If that ghost vehicle doesn't exist, it is the default speed.
    \item Step the CARLA simulation by $\Delta t$. All of $\stackrel{ego}{v}$, $V_{visible}$ are updated appropriately. All $V_{ghost}$ vehicles have their longitudinal positions temporally advanced by their velocity to account for their movement as well.
    \item If a $\stackrel{id}{v} \in V_{visible}$ is now in the ghost regions, assign it internally to $V_{ghost}$ and despawn it in CARLA.
    \item If a $\stackrel{id}{v} \in V_{ghost}$ is now in the visible region, assign it internally to $V_{visible}$ and spawn it in CARLA.
    \item Completely reload the $V_{ghost}$ from $V_{empirical}$ according to the current longitudinal ghost bounds and timestamp.
    \item If $t \ge (t_{0} + t_{max})$, terminate the simulation. Otherwise, return to step 4.

\end{enumerate}

\section{Evaluation and Results}
\begin{figure*}[!h]
    \centering

    \begin{subfigure}{0.98\textwidth}
        \centering
        \caption{Bird's eye view of the cosimulation state at t=10. Green indicates the visible vehicle region, and yellow indicates the ghost cells.}
        \includegraphics[width=0.75\linewidth]{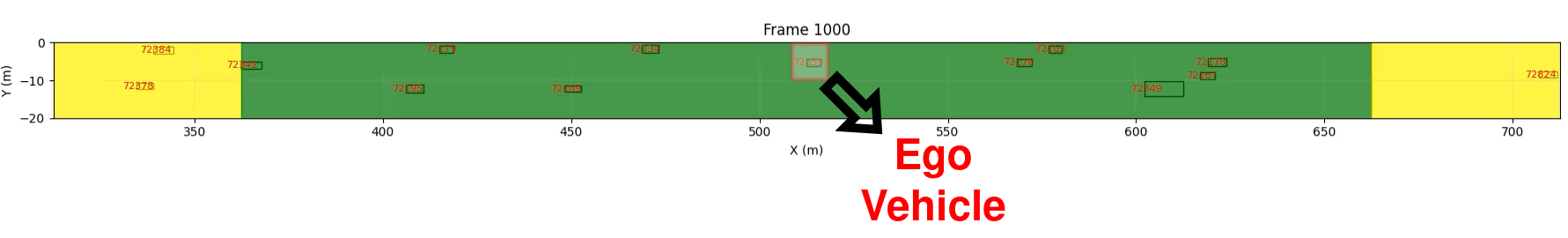}
        \label{fig:scenario_a_bev}
    \end{subfigure}

    \vspace{0.5em} 

    \begin{subfigure}{0.49\textwidth}
        \centering
        \includegraphics[width=0.75\linewidth]{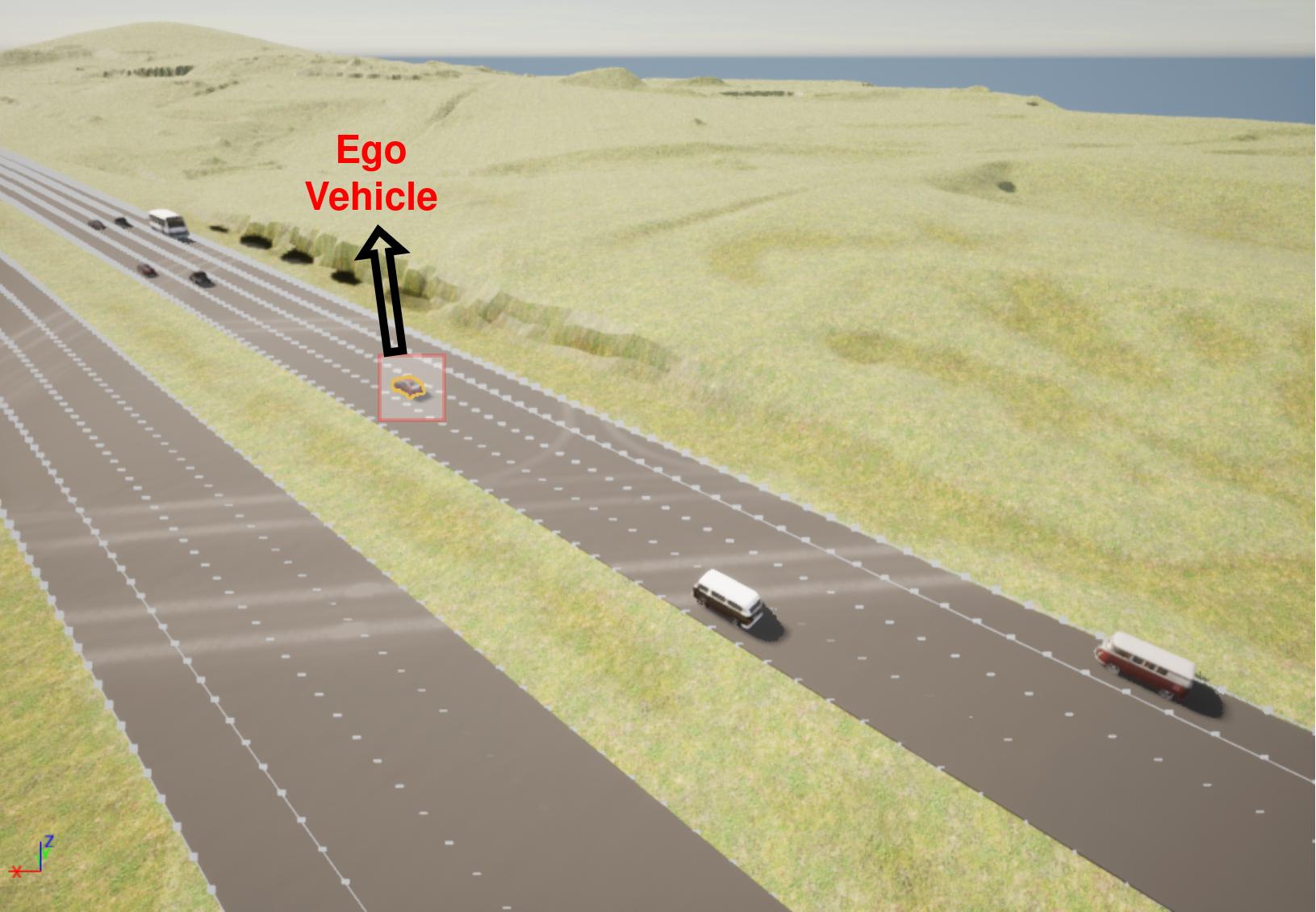}
        \caption{Screenshot of the CARLA simulation at t=10. Ego vehicle highlighted.}
        \label{fig:scenario_a_carla}
    \end{subfigure}
    \hfill
    \begin{subfigure}{0.49\textwidth}
        \centering
        \includegraphics[width=0.75\linewidth]{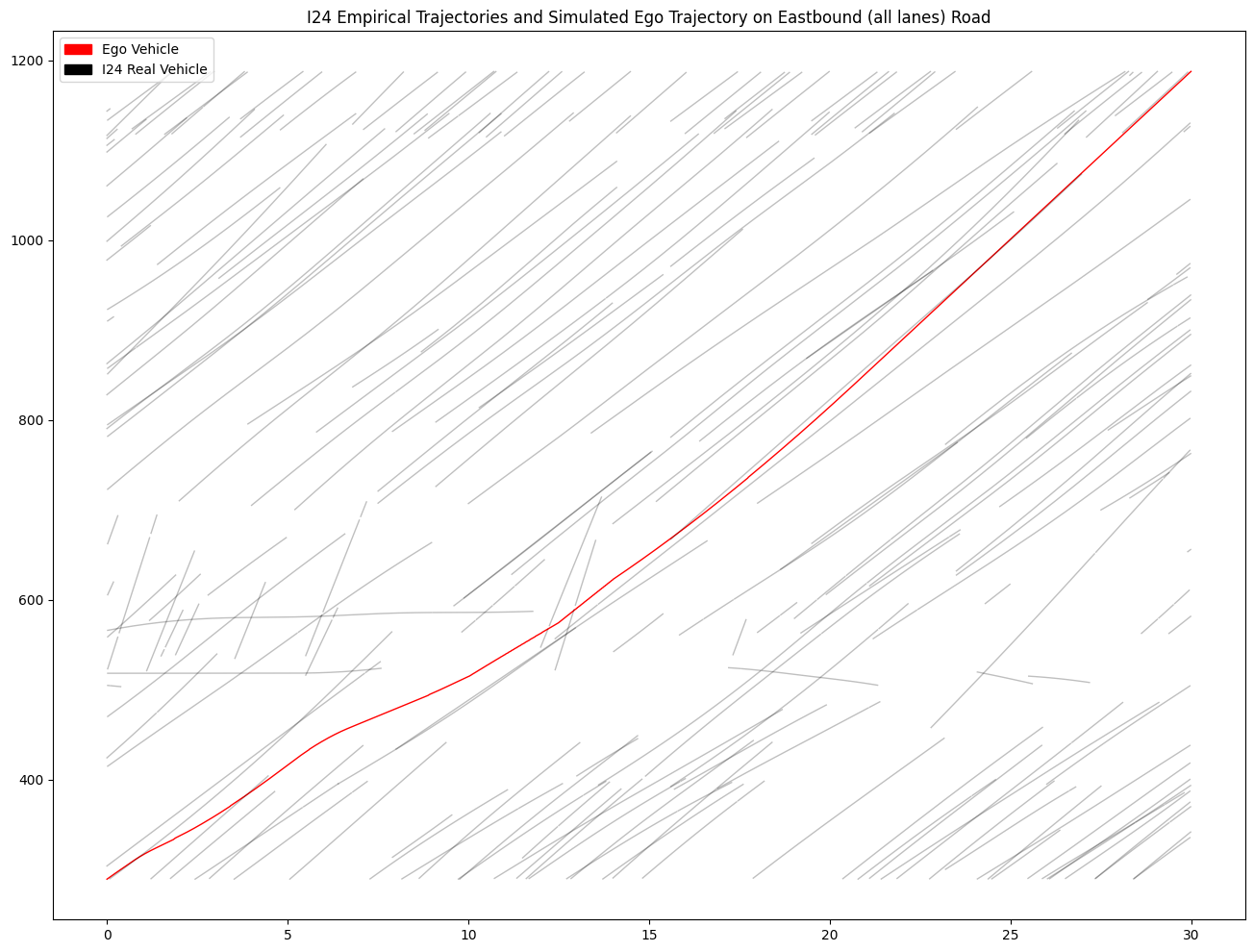}
        \caption{Time-space diagram of the I-24 Empirical Trajectories (all lanes) vs the Simulated Ego Trajectory.}
        \label{fig:scenario_a_empirical}
    \end{subfigure}

    \caption{Scenario A simulation results - a Bird's Eye View of the cosimulation state at (a), a screenshot of CARLA at (b) and a time-space diagram showing the simulated ego vehicle's trajectory vs the I24 empirical trajectories at (c).}
    \label{fig:scenario_a}
\end{figure*}
\begin{figure*}[!h]
    \centering

    \begin{subfigure}{0.98\textwidth}
        \centering
        \caption{Bird's eye view of the cosimulation state at t=10. Green indicates the visible vehicle region, and yellow indicates the ghost cells.}
        \includegraphics[width=0.75\linewidth]{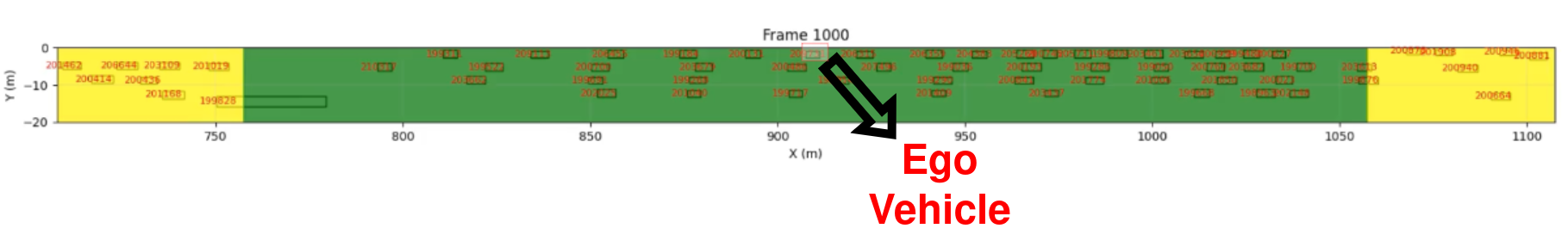}
        \label{fig:scenario_b_bev}
    \end{subfigure}

    \vspace{0.5em} 

    \begin{subfigure}{0.49\textwidth}
        \centering
        \includegraphics[width=0.75\linewidth]{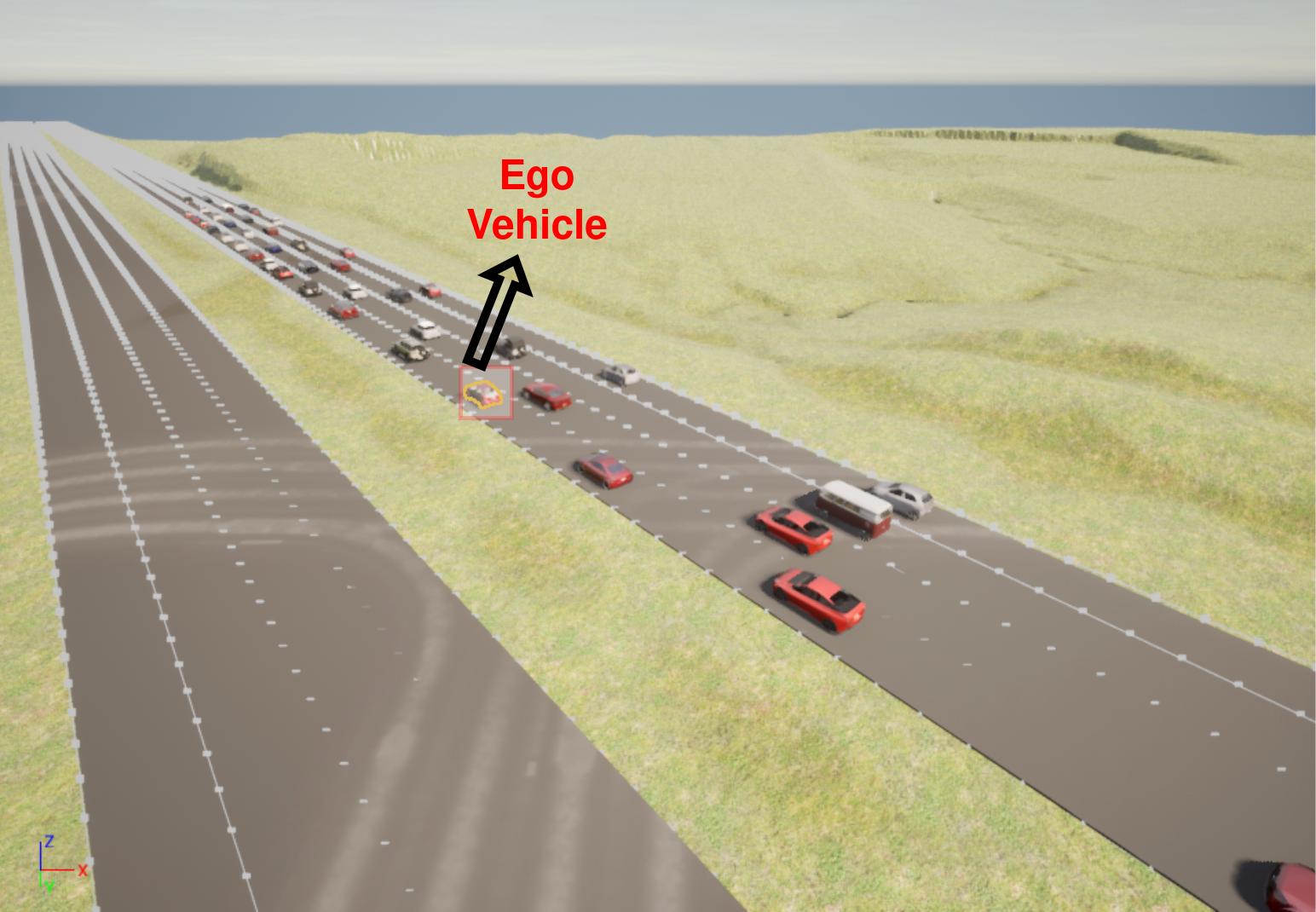}
        \caption{Screenshot of the CARLA simulation at t=10. Ego vehicle highlighted.}
        \label{fig:scenario_b_carla}
    \end{subfigure}
    \hfill
    \begin{subfigure}{0.49\textwidth}
        \centering
        \includegraphics[width=0.75\linewidth]{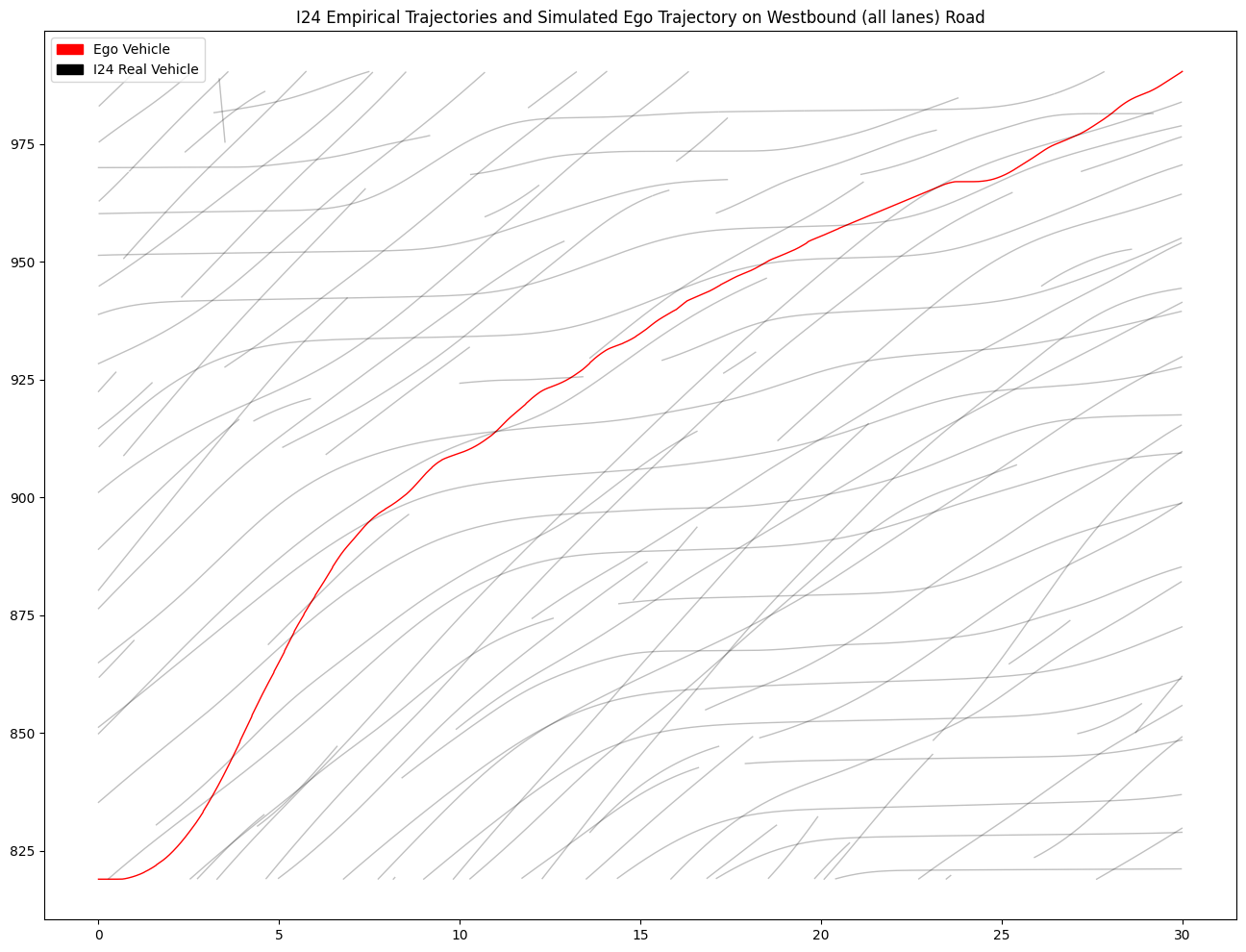}
        \caption{Time-space diagram of the I-24 Empirical Trajectories (all lanes) vs the Simulated Ego Trajectory.}
        \label{fig:scenario_b_empirical}
    \end{subfigure}

    \caption{Scenario B simulation results - a Bird's Eye View of the cosimulation state at (a), a screenshot of CARLA at (b) and a time-space diagram showing the simulated ego vehicle's trajectory vs the I24 empirical trajectories at (c).}
    \label{fig:scenario_b}
\end{figure*}

\subsection{I-24 Empirical Dataset}
Our empirical dataset we use for evaluation is sourced directly from the 2022 I-24 MOTION dataset between the 60.6 and 61.6 mile markers. We select 7 days - 22nd, 23rd, 24th, 25th, 28th, 29th, and 30th of November 2022. Each day lasts between 6 AM and 10 AM CST - which results in 28 hours of potential re-playability.
\subsection{Qualitative Experiments}
\subsubsection{Scenario A}
Scenario A is a 30 second simulation, with the selected ego vehicle on the eastbound road, 2nd lane to the left, at position $x = 300$ meters longitudinally along our 1 mile road, at 6:45:50 AM on November 30th, 2022. The eastbound road is in low-congestion, serving as an excellent baseline. The results of the simulation are shown in Figure \ref{fig:scenario_a}. The results are consistent with the empirical trajectories.

\subsubsection{Scenario B}
Scenario B is a 30 second simulation, with the selected ego vehicle on westbound road, the 1st (HOV) lane to the left, at position $x = 700$ meters longitudinally along our 1 mile road, at 6:45:50 AM on November 30th, 2022. In the empirical data, there is a prominent stop-and-go wave beginning to move through the ego vehicle's position on the HOV lane, with minimal disruption to the 3rd and 4th lanes. Thus, this is an excellent showcase of how the simulation processes stop-and-go waves. The results are shown in Figure \ref{fig:scenario_b}. The simulated ego vehicle's behavior deviates from the empirical data - it does not come to a near-stop. Instead, it merely slows, which is the behavior of the two rightmost lanes in the empirical data. This is due to CARLA's traffic manager inducing vehicles on the HOV lane to lane change into the less congested lanes, improving flow compared to the empirical data. Furthermore, the ego velocity starts at $0 \frac{m}{s}$. This is due to inconsistencies of the efficacy of the impulse-based method for instantaneous acceleration as discussed before.

\subsection{Discussion}
These results indicate the ability of this simulation framework to closely mirror the empirical traffic data ingested from I-24 MOTION, but also illustrate the limitations of the stock behavioral models employed for the ``visible'' vehicles. Since they use CARLA's Traffic Manager, they are not employing IDM. Rather, the behaviors are based on heuristic models from CARLA, which may not reflect the family of behaviors in human drivers. Special lanes in this simulation schema, such as HOV lanes, are treated equally to other lanes, which exacerbates the sim-to-real gap. Thus, in order to improve the behavioral realism of this simulation framework, custom IDM-based agents and lane-changing logic will need to be deployed within it for future work.

\section{Conclusion and Future Work}
In this work, we presented a data-driven traffic microsimulation framework in CARLA that uniquely reconstructs real-world wave dynamics for a local ego vehicle. "Visible" traffic in the ego vehicle's vicinity is able to be directly modeled in CARLA, but still updated and constrained by the global empirical traffic data by the cosimulation module. This will empower future controller testing, autonomy validation, and more. Furthermore, large-scale mesoscopic empirical data allows the fidelity of such microsimulations to be directly evaluated against that data. 

Future work will focus on closing the sim-to-real gap in CARLA's Traffic behavioral models. Our goal is to directly employ such simulations for RL-training of wave-dissipation models, and more. Additional effort will be devoted to analysis of the mathematical framework for convolution of the existing traffic data as ghost cell inputs, to reduce variability of vehicles entering and leaving those cells, which may cause discretized responses in following vehicles. Future work may be able to smooth these disruptions in order to more closely replicate the experience on the open road.

\bibliographystyle{IEEEtran}
\bibliography{ieeeiv2026}

@misc{thiscode,
author={Anonymous},
title={I-24 Motion to CARLA Helpers},
url={https://github.com/AlexOSAdventurer/i24motion_to_carla_helpers},
}

@article{autonomous_ioannou_1993,
	title = {Autonomous intelligent cruise control},
	doi = {10.1109/25.260745},
	author = {Ioannou, Pétros and Chien, Chih‐Ching},
	year = {1993},
	researchRabbitId = {ae35ca1b-311b-4a96-9abe-ee3c93eed16c}
}

@article{multiagent_bhattacharyya_2018,
	title = {Multi-Agent Imitation Learning for Driving Simulation},
	doi = {10.1109/IROS.2018.8593758},
	author = {Bhattacharyya, Raunak P. and Phillips, Derek J. and Wulfe, Blake and Morton, Jeremy and Kuefler, Alex and Kochenderfer, Mykel J.},
	journal = {IEEE/RJS International Conference on Intelligent RObots and Systems},
	year = {2018},
	researchRabbitId = {9646b03b-d9fc-4f86-803e-696cf46f80f9}
}

@article{parsimonious_laval_2014,
	title = {A parsimonious model for the formation of oscillations in car-following models},
	doi = {10.1016/J.TRB.2014.09.004},
	author = {Laval, Jorge and Toth, Christopher and Zhou, Yi},
	journal = {Transportation Research Part B-methodological},
	year = {2014},
	researchRabbitId = {d631fb3b-e797-47a1-a4e2-643ca230d6ba}
}

@article{recurrent_zhou_2017,
	title = {A recurrent neural network based microscopic car following model to predict traffic oscillation},
	doi = {10.1016/J.TRC.2017.08.027},
	author = {Zhou, Mofan and Qu, Xiaobo and Li, Xiaopeng},
	journal = {Transportation Research Part C-emerging Technologies},
	year = {2017},
	researchRabbitId = {ec6129dd-81d6-4e93-99b0-27bb79f9be4c}
}

@article{neural_panwai_2007,
	title = {Neural Agent Car-Following Models},
	doi = {10.1109/TITS.2006.884616},
	author = {Panwai, Sakda and Dia, Hussein},
	journal = {IEEE Transactions on Intelligent Transportation Systems},
	year = {2007},
	researchRabbitId = {2ec24537-c61f-44e1-a5f6-63b3b40799b0}
}

@article{enhanced_sharath_2020,
	title = {Enhanced intelligent driver model for two-dimensional motion planning in mixed traffic},
	doi = {10.1016/J.TRC.2020.102780},
	author = {Sharath, M.N. and Velaga, Nagendra R.},
	journal = {Transportation Research Part C-emerging Technologies},
	year = {2020},
	researchRabbitId = {0b8d847f-7cee-47ea-a772-5983c1446887}
}

@article{survey_chao_2020,
	title = {A Survey on Visual Traffic Simulation: Models, Evaluations, and Applications in Autonomous Driving},
	doi = {10.1111/CGF.13803},
	author = {Chao, Qianwen and Bi, Huikun and Li, Weizi and Mao, Tianlu and Wang, Zhaoqi and Lin, Ming C. and Deng, Zhigang},
	journal = {Computer graphics forum (Print)},
	year = {2020},
	researchRabbitId = {6a334c99-f208-45e2-9276-8fe58e1f27c8}
}

@article{highd_krajewski_2018,
	title = {The highD Dataset: A Drone Dataset of Naturalistic Vehicle Trajectories on German Highways for Validation of Highly Automated Driving Systems},
	doi = {10.1109/ITSC.2018.8569552},
	author = {Krajewski, Robert and Bock, Julian and Kloeker, Laurent and Eckstein, Lutz},
	journal = {International Conference on Intelligent Transportation Systems},
	year = {2018},
	researchRabbitId = {290b995d-1a33-4c69-afd5-930e5e4dd2f8}
}

@article{scalable_ji_2024,
	title = {Scalable analysis of stop-and-go waves},
	doi = {10.48550/ARXIV.2409.00326},
	author = {Ji, Junyi and Gloudemans, Derek and Wang, Yanbing and Zachár, Gergely and Barbour, William and Sprinkle, Jonathan and Piccoli, Benedetto and Work, Daniel B.},
	journal = {arXiv.org},
	year = {2024},
	researchRabbitId = {c477873b-264a-4425-a30d-6015db75c2de}
}

@article{traffic_lee_2025,
	title = {Traffic Control via Connected and Automated Vehicles (CAVs): An Open-Road Field Experiment with 100 CAVs},
	doi = {10.1109/MCS.2024.3498552},
	author = {Lee, Jonathan W. and others},
    overflowauthors = {Wang, Han and Jang, Kathy and Lichtlé, Nathan and Hayat, Amaury and Bunting, Matt and Alanqary, Arwa and Barbour, William and Fu, Zhe and Gong, Xiaoqian and Gunter, George and Hornstein, Sharon and Kreidieh, Abdul Rahman and Nice, Matthew and Richardson, William A. and Shah, Adit and Vinitsky, Eugene and Wu, Fangyu and Xiang, Shengquan and Almatrudi, Sulaiman and Althukair, Fahd A. and Bhadani, R. and Carpio, Joy and Chekroun, Raphael and Cheng, Eric and Chiri, M. and Chou, Fang-Chieh and Delorenzo, Ryan and Gibson, Marsalis T. and Gloudemans, Derek and Gollakota, Anish and Ji, Junyi and Keimer, Alexander and Khoudari, Nour and Mahmood, Malaika and Mahmood, Mikail and Matin, Hossein Nick Zinat and McQuade, Sean T. and Ramadan, Rabie and Urieli, Daniel and Wang, Xia and Wang, Yanbing and Xu, Rita and Yao, Mengsha and You, Yiling and Zachár, G. and Zhao, Yibo and Ameli, Mostafa and Baig, Mirza Najamuddin and Bhaskaran, Sarah and Butts, Kenneth and Gowda, Manasi and Janssen, Caroline and Lee, John and Pedersen, Liam and Wagner, Riley and Zhang, Zimo and Zhou, Chang and Work, Daniel B. and Seibold, Benjamin and Sprinkle, Jonathan and Piccoli, Benedetto and Monache, M. D. and Bayen, Alexandre M.},
	journal = {IEEE Control Systems},
	year = {2025},
	researchRabbitId = {4783d927-5968-4a80-b6f5-a248fd371e39}
}

@article{gloudemans2023i24motion,
title = {I-24 MOTION: An instrument for freeway traffic science},
journal = {Transportation Research Part C: Emerging Technologies},
volume = {155},
pages = {104311},
year = {2023},
issn = {0968-090X},
doi = {https://doi.org/10.1016/j.trc.2023.104311},
url = {https://www.sciencedirect.com/science/article/pii/S0968090X23003005},
author = {Derek Gloudemans and Yanbing Wang and Junyi Ji and Gergely Zachár and William Barbour and Eric Hall and Meredith Cebelak and Lee Smith and Daniel B. Work}
}

@article{ros_elmadani_2021,
	title = {From CAN to ROS: A Monitoring and Data Recording Bridge},
	doi = {10.1145/3459609.3460531},
	author = {Elmadani, Safwan and Nice, Matthew and Bunting, Matt and Sprinkle, Jonathan and Bhadani, Rahul},
	journal = {Proceedings of the Workshop on Data-Driven and Intelligent Cyber-Physical Systems},
	year = {2021},
	researchRabbitId = {8d0cb021-7213-42e3-a262-c648254eb7b1}
}

@article{libpanda_bunting_2021,
	title = {Libpanda: A High Performance Library for Vehicle Data Collection},
	doi = {10.1145/3459609.3460529},
	author = {Bunting, Matt and Bhadani, Rahul and Sprinkle, Jonathan},
	journal = {Proceedings of the Workshop on Data-Driven and Intelligent Cyber-Physical Systems},
	year = {2021},
	researchRabbitId = {c368442b-bcda-4c70-9023-ebf968026c60}
}

@article{behavioural_gipps_1981,
	title = {A behavioural car-following model for computer simulation},
	doi = {10.1016/0191-2615(81)90037-0},
	author = {Gipps, P G},
	journal = {Transportation Research Part B-methodological},
	year = {1981},
	researchRabbitId = {c1971f0e-92b8-4f5c-a7e8-58b70348464d}
}

@article{dynamical_bando_1995,
	title = {Dynamical model of traffic congestion and numerical simulation.},
	doi = {10.1103/PHYSREVE.51.1035},
	author = {Bando, Masako and Hasebe, Katsuya and Nakayama, A. and Shibata, Akira and Sugiyama, Yūki},
	journal = {Physical review. E, Statistical physics, plasmas, fluids, and related interdisciplinary topics},
	year = {1995},
	pubmedId = {https://pubmed.ncbi.nlm.nih.gov/9962746},
	researchRabbitId = {22ba5c99-4fe0-4c7f-8a5e-be5abc851969}
}

@article{intelligent_treiber_2017,
	title = {The Intelligent Driver Model with stochasticity--New insights into traffic flow oscillations},
	doi = {10.1016/J.TRB.2017.08.012},
	author = {Treiber, Martin and Kesting, Arne},
	journal = {Transportation Research Part B: Methodological},
	year = {2017},
	researchRabbitId = {74354246-6da9-4507-b9a7-4a2937ccce35}
}

@article{survey_di_2021,
	title = {A survey on autonomous vehicle control in the era of mixed-autonomy: From physics-based to AI-guided driving policy learning},
	doi = {10.1016/J.TRC.2021.103008},
	author = {Di, Xuan and Shi, Rongye},
	journal = {Transportation Research Part C: Emerging Technologies},
	year = {2021},
	researchRabbitId = {8c19266b-a399-420a-ac72-23361a93d96c}
}

@article{traffic_treiber_2012,
	title = {Traffic Flow Dynamics},
	doi = {10.1007/978-3-642-32460-4},
	author = {Treiber, Martin and Kesting, Arne},
	year = {2012},
	researchRabbitId = {a15c0224-e934-485e-a50a-8f0aea6d4e19}
}

@article{congested_treiber_2000,
	title = {Congested traffic states in empirical observations and microscopic simulations},
	doi = {10.1103/PHYSREVE.62.1805},
	author = {Treiber, Martin and Hennecke, Ansgar and Helbing, Dirk},
	journal = {Physical Review E},
	year = {2000},
	pubmedId = {https://pubmed.ncbi.nlm.nih.gov/11088643},
	researchRabbitId = {8e93dcbd-71db-44fe-aa6c-319f0551c8a8}
}

@article{microscopic_lopez_2018,
	title = {Microscopic Traffic Simulation using SUMO},
	doi = {10.1109/ITSC.2018.8569938},
	author = {Lopez, Pablo Alvarez and Wiebner, Evamarie and Behrisch, Michael and Bieker-Walz, Laura and Erdmann, Jakob and Flotteröd, Yun-Pang and Hilbrich, Robert and Lücken, Leonhard and Rummel, Johannes and Wagner, Péter},
	journal = {International Conference on Intelligent Transportation Systems},
	year = {2018},
	researchRabbitId = {6dc80c25-eb08-4ae8-8ff0-132286b78698}
}

@article{benchmarks_vinitsky_2018,
	title = {Benchmarks for reinforcement learning in mixed-autonomy traffic},
	author = {Vinitsky, Eugene and Kreidieh, Aboudy and Flem, Luc Le and Kheterpal, Nishant and Jang, Kathy and Wu, Cathy and Wu, Fangyu and Liaw, Richard and Liang, Eric},
	journal = {Conference on Robot Learning},
	year = {2018},
	researchRabbitId = {c0f49ca5-76dd-4b4f-8992-ed5d15e70a74}
}

@article{flow_wu_2017,
	title = {Flow: Architecture and Benchmarking for Reinforcement Learning in Traffic Control.},
	author = {Wu, Cathy and Kreidieh, Aboudy and Parvate, Kanaad and Vinitsky, Eugene and Bayen, Alexandre M.},
	journal = {arXiv.org},
	year = {2017},
	researchRabbitId = {8ce221af-d2ab-4b9f-bde1-4985d36cdda3}
}

@article{traffic_sugiyama_2008,
	title = {Traffic jams without bottlenecks--experimental evidence for the physical mechanism of the formation of a jam},
	doi = {10.1088/1367-2630/10/3/033001},
	author = {Sugiyama, Yūki and Fukui, Minoru and Kikuchi, Macoto and Hasebe, Katsuya and Nakayama, Akihiro and Nishinari, Katsuhiro and Tadaki, Shin-ichi and Yukawa, Satoshi},
	year = {2008},
	researchRabbitId = {7a272164-74c2-4cf9-a176-264900011b5b}
}

@article{i24_gloudemans_2023,
	title = {I-24 MOTION: An instrument for freeway traffic science},
	doi = {10.48550/ARXIV.2301.11198},
	author = {Gloudemans, Derek and Wang, Yanbing and Ji, Junyi and Zachár, Gergely and Barbour, Will and Work, Daniel B.},
	journal = {arXiv (Cornell University)},
	year = {2023},
	researchRabbitId = {7a51e1d1-42e1-40d2-9686-1b5bfbcf824f}
}

@article{flow_wu_2021,
	title = {Flow: A Modular Learning Framework for Mixed Autonomy Traffic},
	doi = {10.1109/TRO.2021.3087314},
	author = {Wu, Cathy and Kreidieh, Abdul Rahman and Parvate, Kanaad and Vinitsky, Eugene and Bayen, Alexandre M.},
	journal = {IEEE Transactions on robotics},
	year = {2021},
	researchRabbitId = {8b596ade-94af-4558-9d70-7375fd669094}
}

@article{carla_dosovitskiy_2017,
	title = {CARLA: An Open Urban Driving Simulator},
	author = {Dosovitskiy, Alexey and Ros, German and Codevilla, Felipe and Lopez, Antonio M. and Koltun, Vladlen},
	journal = {arXiv: Learning},
	year = {2017},
	researchRabbitId = {b2e8876a-ab44-45d7-8b9e-81209d71b67f}
}

@article{traffic_zehe_2015,
	title = {Traffic Simulation Performance Optimization through Multi-Resolution Modeling of Road Segments},
	doi = {10.1145/2769458.2769475},
	author = {Zehe, Daniel and Grotzky, David and Aydt, Heiko and Cai, Wentong and Knoll, Alois},
	journal = {SIGSIM-PADS},
	year = {2015},
	researchRabbitId = {870c1957-f40e-411e-98f1-6d78066a5eef}
}

@article{safety_xu_2025,
	title = {Safety validation for connected autonomous vehicles using large-scale testing tracks in high-fidelity simulation environment},
	doi = {10.1016/J.AAP.2025.108011},
	author = {Xu, Zheng and Wang, Xiaomeng and Wang, Xuesong and Zheng, Nan},
	journal = {Accident Analysis \& Prevention},
	year = {2025},
	pubmedId = {https://pubmed.ncbi.nlm.nih.gov/40107085},
	researchRabbitId = {f8e8b75e-7433-45ef-88c0-413911a2ac4f}
}

@article{carla_osiski_2020,
	title = {CARLA Real Traffic Scenarios - novel training ground and benchmark for autonomous driving},
	author = {Osinski, Blażej and Milos, Piotr and Jakubowski, Adam and Ziecina, Paweł and Martyniak, Michal and Galias, Christopher and Breuer, Antonia and Homoceanu, Silviu and Michalewski, Henryk},
	journal = {arXiv.org},
	year = {2020},
	researchRabbitId = {bff3d61c-c12a-4803-a92d-2ba8c6479433}
}

@article{commercially_gunter_2019,
	title = {Are commercially implemented adaptive cruise control systems string
  stable?},
	doi = {10.48550/ARXIV.1905.02108},
	author = {Gunter, George and Gloudemans, Derek and Stern, Raphael and McQuade, Sean T. and Bhadani, Rahul and Bunting, Matt and Monache, Maria Laura Delle and Lysecky, Roman and Seibold, Benjamin and Sprinkle, Jonathan and Piccoli, Benedetto and Work, Daniel B.},
	journal = {arXiv (Cornell University)},
	year = {2019},
	researchRabbitId = {e0881f50-d8d1-4ed6-a507-29d0793c9c5c}
}

\end{document}